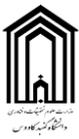
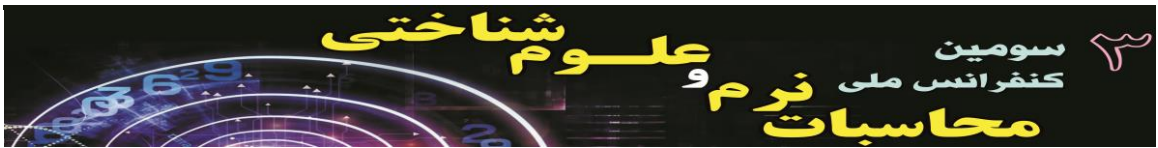

# A Comprehensive Machine Learning Framework for Heart Disease Prediction: Performance Evaluation and Future Perspectives

**Ali Azimi Lamir [1], Shiva Razzagzadeh [2], Zeynab Rezaei [3]**

[1] Department of Computer Engineering, Ardabil Branch, Islamic Azad University, Ardabil, Iran;
Aliazimi71930@gmail.com

[2] Department of Computer Engineering, Ardabil Branch, Islamic Azad University, Ardabil, Iran;
Shiva.razzaghzadeh@gmail.com

[3] Department of Computer Engineering, Ardabil Branch, Islamic Azad University, Ardabil, Iran;
Rezaii57@yahoo.com

**ABSTRACT**

This study presents a machine learning-based framework for heart disease prediction using the heart-disease dataset, comprising 303 samples with 14 features. The methodology involves data preprocessing, model training, and evaluation using three classifiers: Logistic Regression, K-Nearest Neighbors (KNN), and Random Forest. Hyperparameter tuning with GridSearchCV and RandomizedSearchCV was employed to enhance model performance. The Random Forest classifier outperformed other models, achieving an accuracy of 91% and an F1-score of 0.89. Evaluation metrics, including precision, recall, and confusion matrix, revealed balanced performance across classes. The proposed model demonstrates strong potential for aiding clinical decision-making by effectively predicting heart disease. Limitations such as dataset size and generalizability underscore the need for future studies using larger and more diverse datasets. This work highlights the utility of machine learning in healthcare, offering insights for further advancements in predictive diagnostics.

## 1. INTRODUCTION

Heart disease, a common and significant medical condition, has a deep interconnection with the wider healthcare sector [1]. Cardiovascular disease is a major factor in morbidity and mortality globally, highlighting the essential function of healthcare systems in addressing urgent health issues. Heart disease encompasses several disorders affecting the heart and blood vessels, including heart failure, coronary artery disease, arrhythmias, and valvular anomalies. Healthcare experts are essential in diagnosing cardiac disease through medical examinations, advanced imaging techniques, and diagnostic procedures [2]. The management of the illness requires a multifaceted strategy that includes lifestyle modifications, pharmacological treatments, and surgical interventions, emphasising the necessity for interdisciplinary collaboration among healthcare practitioners. Healthcare initiatives that emphasise the promotion of heart-healthy practices, the improvement of awareness, and the support for early diagnosis are essential in reducing the incidence of heart disease and improving cardiovascular health on a larger scale. Investigating and managing cardiovascular disease illustrates the intricate relationship between healthcare and societal welfare.

Due to the intricate nature of heart disease, it necessitates careful management. Neglecting to do so may inflict harm on the heart or lead to premature death. Heart attacks manifest abruptly in over 40% of instances. Notwithstanding optimal medical intervention, such occurrences frequently result in fatality and are exceedingly grave, rendering life preservation unattainable. Early prediction of CHF can facilitate the discovery of a cure and preserve several lives. Diagnostic procedures for CHF include X-ray, B.N.P. (brain natriuretic peptide), echocardiography, and cardiac catheterisation. Manual techniques for detecting congestive heart failure (CHF) encompass the detection of jugular vein distention in the neck, peripheral oedema, systemic congestion, and potentially hepatic cirrhosis. The quality of service (QoS) in the healthcare sector, which ensures precise and timely disease diagnosis and proficient patient care, is facing significant problems. A specifically educated artificial intelligence (AI) model can assist medical personnel in diagnosis. This can expedite the diagnostic process and alleviate their workload. This can also conserve valuable time for medical experts. In this context, machine learning (ML) algorithms serve as useful and trustworthy tools for recognising and classifying patients with and without congestive heart failure (CHF). Consequently, researchers have developed a diverse array of automated diagnostic systems utilising machine learning and data mining methodologies [5,6]. The existing methodologies exhibit certain constraints, including the substantial data generated by CT and MRI scans and spatial complexity, which complicates the development of a solution to differentiate between heart disease and stroke. The prediction of heart disease is regarded as difficult; nonetheless, the advancement of machine learning algorithms has rendered it a significant topic of study. The utilisation of AI enables both patients and physicians to preserve lives by initiating therapy promptly using electronic health records (EHRs), which are advantageous in contemporary times for clinical and research applications, enhancing productivity and efficacy in healthcare. To enhance the precision and reliability of the prediction system, there is a pressing need to improve accuracy.

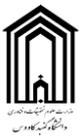
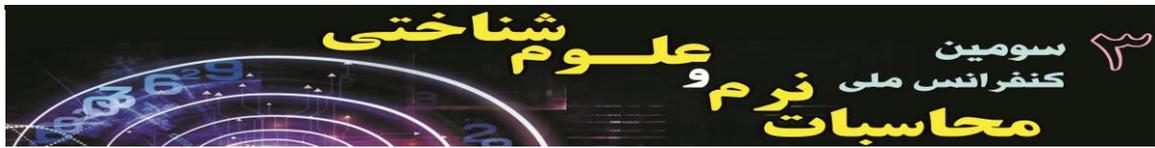

Various machine learning techniques have been utilised to reduce prediction error and align with actual outcomes, thereby elucidating the non-linear relationships among diverse parameters and complexities. Clinical research and data mining methodologies are employed to identify multiple forms of anabolic syndromes. Classification and data mining play an essential role in data analysis and the prediction of cardiac disease. Additionally, we have seen the application of decision trees to detect occurrences associated with heart disease [7]. Nonetheless, in the context of machine learning, imbalanced and outlier data may arise and impact the efficacy of the predictive model. Previous studies have demonstrated that outlier data can be detected and removed through density-based spatial clustering of applications with noise methods [8]. Additionally, to equilibrate data distribution, a hybrid synthetic minority over-sampling technique-edited nearest neighbour (SMOTE-ENN) is employed to markedly enhance the efficacy of prediction models [9].

This study seeks to investigate the use of machine learning models, namely Logistic Regression, KNN, and Random Forest, for predicting cardiac disease using a publically accessible dataset. We concentrate on assessing the efficacy of these models regarding accuracy, precision, recall, and F1-score. Additionally, we examine the advantages of hyperparameter adjustment to enhance model performance and increase forecast accuracy. This study aims to enhance the existing research on machine learning in healthcare and illustrate the practical applicability of these models in facilitating the early diagnosis of heart disease.

This work intends to enhance the effectiveness of heart disease prediction through machine learning techniques while simultaneously emphasising the limitations of existing methods and the necessity for larger, more diverse datasets. The results herein will yield significant insights into the advantages and disadvantages of various machine learning models, establishing a basis for subsequent research in this vital domain of healthcare.

## 2. Related Work

The non-linear Cleveland heart disease dataset was employed in [10] to predict heart disease by the application of random forests with slight adjustments. All attributes with incomplete data were identified, and the median of each attribute was utilised to substitute the missing values. The heart disease dataset was sanitised by eliminating any missing values. They attained more precise heart disease predictions by utilising the RF classification method when the attributes were clearly delineated. The accuracy of the random forest model was validated using 10-fold cross-validation after training on 303 data examples. The proposed model surpassed the other models by attaining optimal accuracy.

Gjoreski et al. presented a cardiac sound-based method for detecting congestive heart failure in [11]. They focused on analysing cardiac sound recordings to ascertain the status of congestive heart failure (CHF). The methodology integrated the conventional machine learning strategy with the end-to-end deep learning approach. A spectro-temporal representation of the signal served as input for the deep learning algorithm, while standard machine learning utilised expert features for training. A total of 947 publicly accessible individual recordings and one CHF dataset were collected, and the proposed approach was assessed. The machine learning models were developed utilising 15 features to distinguish between phases of congestive heart failure.

Plati et al. introduced a model employing various machine learning classifiers, including Bayes network (BN), decision tree, support vector machine (SVM), logistic model tree (LMT), random forest (RF), k-nearest neighbours (KNN), naive Bayes (NB), and rotation forest (ROT), to detect heart failure in 422 participants, utilising 10-fold cross-validation for assessment [12]. The HF sample consisted of 73 female and 154 male respondents, while the non-HF dataset comprised 106 male and 151 female subjects. LMT and ROT outperformed all other classifiers in the comparison.

Gjoreski et al. proposed a technique including the stacking of various machine learning classifiers to identify congestive heart failure from cardiac sounds [13]. A professional digital stethoscope was employed to record the sounds. They utilised 152 unique heart sounds from a total of 122 individuals. During the segment-based machine learning phase, tests were performed utilising combinations of several model types, encompassing both individual techniques and a composite of seven techniques—Naive Bayes, Bagging, Random Forest, Support Vector Machine, K-Nearest Neighbours, Boosting, and J48. For the assessment, these models utilised the leave-one-subject-out (LOSO) cross-validation technique. The experimental approach reached an accuracy of 96%, yielding promising results.

Aljaaf et al. developed a heart failure risk assessment prediction model utilising the C4.5 classifier over many tiers, categorising heart failure into five risk levels: high-risk, low-risk, extreme-risk, moderate-risk, and no-risk [14]. This research utilised cardiac disease data from the Cleveland Clinic Foundation. Three additional factors—physical activity, smoking, and obesity—that substantially elevate the risk of heart failure were incorporated into the dataset to enhance its features in the study. Ten-fold cross-validation was utilised to assess performance. The prediction model has superior performance with 95.5% specificity, 86.5% sensitivity, and 86.53% accuracy compared to many other models.

## 3. Material and Methods

The proposed method for heart disease classification involves a comprehensive end-to-end pipeline that integrates exploratory data analysis, machine learning model development, and hyperparameter tuning. The dataset utilized in this study, heart-disease.csv, comprises 303 samples with 14 features related to patient demographics, clinical measurements, and disease status. The data is divided into training and testing sets using an 80-20 split ratio via train_test_split to ensure robust model evaluation.

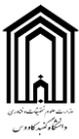
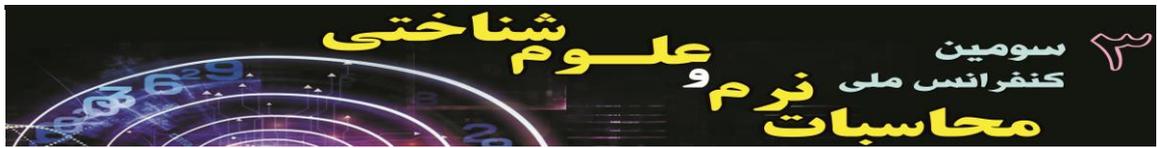

Three machine learning models were implemented: Logistic Regression, K-Nearest Neighbors (KNN), and Random Forest. To optimize model performance, hyperparameter tuning was conducted using GridSearchCV and RandomizedSearchCV. Evaluation metrics include precision, recall, F1-score, and confusion matrix, ensuring a detailed analysis of model performance. Visualization of results using libraries like Matplotlib and Seaborn further aids in interpreting classification outcomes. This systematic approach aims to identify the most accurate and interpretable model for predicting heart disease, leveraging the strengths of ensemble learning and hyperparameter optimization.

### 3.1. Dataset

The dataset used in this study, heart-disease.csv, contains 303 instances with 14 features. These features include patient demographics (e.g., age, sex), clinical measurements (e.g., cholesterol level, maximum heart rate achieved), and the target variable indicating the presence or absence of heart disease. The dataset is publicly available and commonly used for benchmarking machine learning models for heart disease prediction. The following are the features we'll use to predict our target variable (heart disease or no heart disease).

**Tabel 1. Information Dataset.**

| Feature | Description | Example Values |
|---|---|---|
| **age** | Age in years | 29, 45, 60 |
| **sex** | 1 = male; 0 = female | 0, 1 |
| **cp** | Chest pain type | 0: Typical angina (chest pain), 1: Atypical angina (chest pain not related to heart), 2: Non-anginal pain (typically esophageal spasms (non heart related), 3: Asymptomatic (chest pain not showing signs of disease) |
| **trestbps** | Resting blood pressure (in mm Hg on admission to the hospital) | 120, 140, 150 |
| **chol** | Serum cholesterol in mg/dl | 180, 220, 250 |
| **fbs** | Fasting blood sugar > 120 mg/dl (1 = true; 0 = false) | 0, 1 |
| **restecg** | Resting electrocardiographic results | 0: Nothing to note, 1: ST-T Wave abnormality, 2: Left ventricular hypertrophy |
| **thalach** | Maximum heart rate achieved | 160, 180, 190 |
| **exang** | Exercise induced angina (1 = yes; 0 = no) | 0, 1 |
| **oldpeak** | ST depression (heart potentially not getting enough oxygen) induced by exercise relative to rest | 0.5, 1.0, 2.0 |
| **slope** | The slope of the peak exercise ST segment | 0: Upsloping, 1: Flatsloping, 2: Downsloping |
| **ca** | Number of major vessels (0-3) colored by fluoroscopy | 0, 1, 2, 3 |
| **thal** | Thalium stress result | 1: Normal, 3: Normal, 6: Fixed defect, 7: Reversible defect |
| **target** | Have disease or not (1 = yes; 0 = no) | 0, 1 |

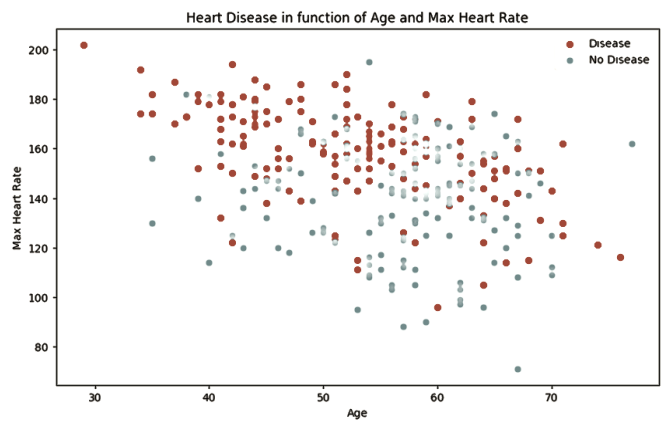

**Figure 1. Association Between Age, Heart Rate, and Disease Status**

The scatter plot illustrates the correlation between age, maximum heart rate, and the presence of heart disease. Each data point represents an individual, with the x-axis indicating age and the y-axis representing maximum heart rate. The color of the data point signifies the presence or absence of heart disease. The graph suggests that there might be a trend or pattern between these variables, indicating that older individuals with lower maximum heart rates may be at a higher risk of developing heart disease. However, further statistical analysis is required to confirm any significant relationships and draw definitive conclusions.

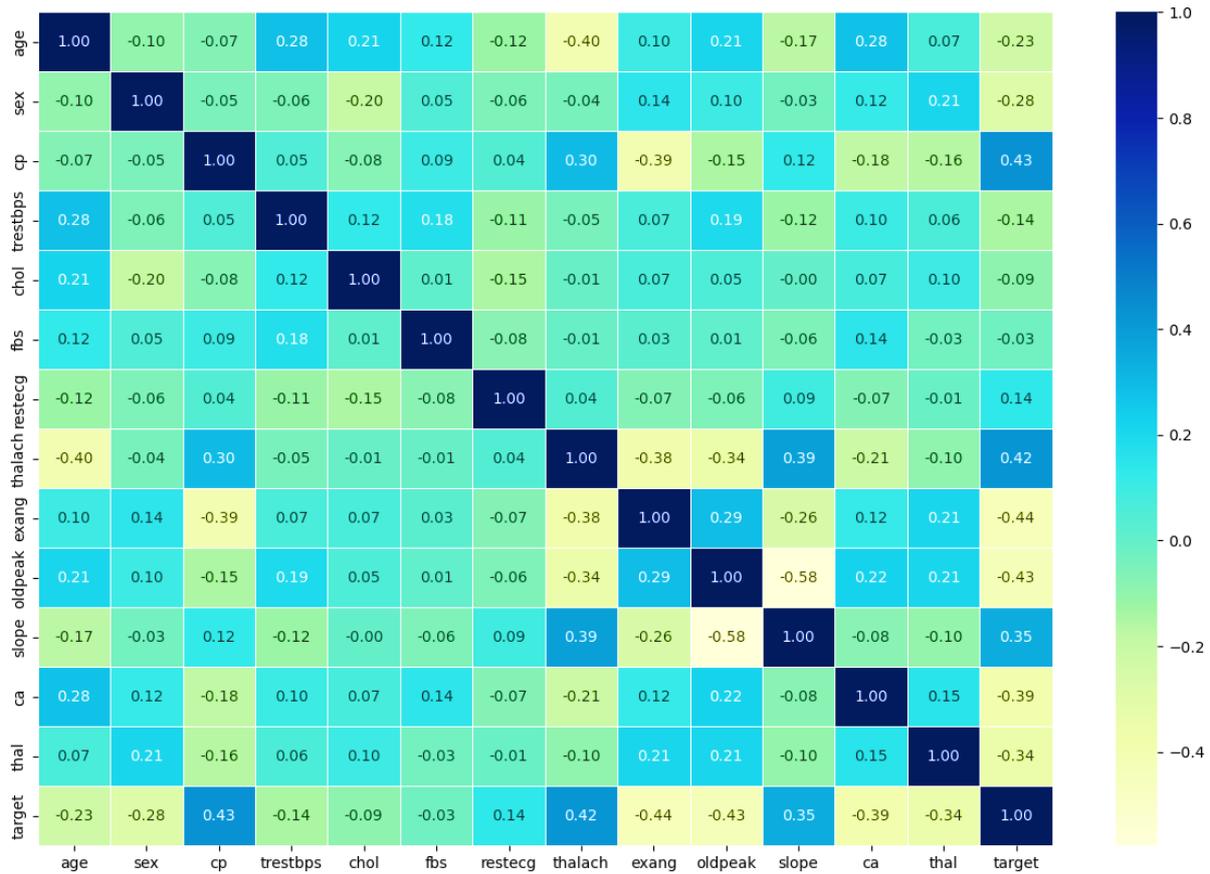

**Figure 2. scatter plot visualizes the relationship between age, maximum heart rate, and the presence of heart disease.**

The scatter plot illustrates the correlation between age, maximum heart rate, and the presence of heart disease. Each data point represents an individual, with the x-axis indicating age and the y-axis representing maximum heart rate. The color of the data point signifies the presence or absence of heart disease. The graph suggests that there might be a trend or pattern between these variables, indicating that older individuals with lower maximum heart rates may be at a higher risk of developing heart disease. However, further statistical analysis is required to confirm any significant relationships and draw definitive conclusions.

### 3.2. Data Preprocessing

Initial preprocessing steps include handling missing values, scaling numerical features using standardization, and encoding categorical variables into numeric form. The dataset was split into training (80%) and testing (20%) subsets using train_test_split. To prevent data leakage, feature scaling and transformations were applied only to the training set before being applied to the test set. Exploratory Data Analysis (EDA) identified feature correlations to guide model development.

### 3.3. Model Architecture

The proposed method for heart disease classification involves a comprehensive end-to-end pipeline that integrates exploratory data analysis, machine learning model development, and hyperparameter tuning. The dataset utilized in this study, heart-disease.csv, comprises 303 samples with 14 features related to patient demographics, clinical measurements, and disease status. The data is divided into training and testing sets using an 80-20 split ratio via train_test_split to ensure robust model evaluation.

Three machine learning models were implemented: Logistic Regression, K-Nearest Neighbors (KNN), and Random Forest. To optimize model performance, hyperparameter tuning was conducted using GridSearchCV and RandomizedSearchCV. Evaluation metrics include precision, recall, F1-score, and confusion matrix, ensuring a detailed analysis of model performance. Visualization of results using libraries like Matplotlib and Seaborn further aids in interpreting classification outcomes. This systematic approach aims to identify the most accurate and interpretable model for predicting heart disease, leveraging the strengths of ensemble learning and hyperparameter optimization.

Three machine learning models were employed for heart disease prediction:

- Logistic Regression: A linear model suitable for binary classification tasks.
- K-Nearest Neighbors (KNN): A non-parametric model relying on feature similarity for classification.
- Random Forest: An ensemble model combining multiple decision trees to enhance accuracy and reduce overfitting.
- GridSearchCV and RandomizedSearchCV were used for hyperparameter tuning, optimizing parameters such as the number of neighbors for KNN and the number of trees for Random Forest.

### 3.4. Model Training and Evaluation
Models were trained on the training dataset, and evaluation was performed on the unseen test dataset. Evaluation metrics included precision, recall, F1-score, and confusion matrix. These metrics provide a comprehensive understanding of model performance, particularly for imbalanced datasets.

### 3.5. Evaluation and Results
In this section, we present the evaluation results of the machine learning models used for heart disease prediction, focusing on the performance of Logistic Regression, K-Nearest Neighbors (KNN), and Random Forest. The models were assessed using several key metrics, including accuracy, precision, recall, and F1-score, to provide a comprehensive view of their effectiveness. Additionally, hyperparameter tuning was conducted to optimize model performance, and the results are analyzed in terms of their ability to distinguish between patients with and without heart disease. The following subsections detail the performance of each model, with a particular emphasis on the strengths and weaknesses observed during evaluation.

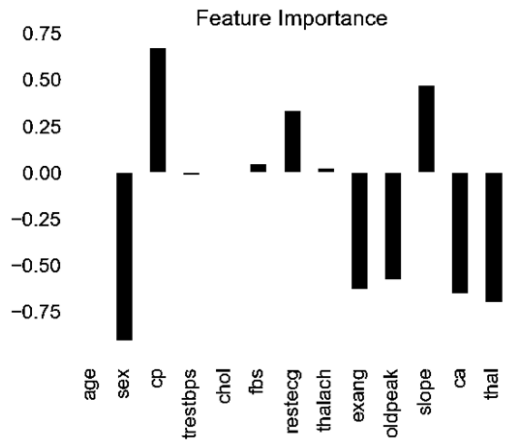

**Figure 3. Feature Importance: A Ranking of Predictive Factors**

The bar chart illustrates the relative importance of each feature in a predictive model. The height and direction of each bar represent the feature's contribution to the model's outcome. Positive values indicate a positive correlation with the target variable, while negative values indicate a negative correlation. For instance, 'cp' and 'slope' have a strong positive impact on the model's predictions, suggesting that these features are highly influential in determining the outcome. Conversely, 'age' has a significant negative impact, implying that as age increases, the outcome is likely to decrease.

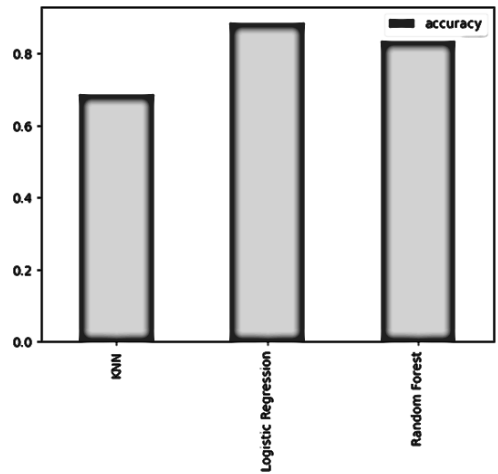

**Figure 4. Classification Accuracy Across Different Algorithms**

The bar chart compares the accuracy of three machine learning models: KNN, Logistic Regression, and Random Forest. Logistic Regression achieved the highest accuracy, followed by Random Forest. KNN performed the worst among the three models. This suggests that Logistic Regression is the most suitable model for this particular dataset. However, it's important to consider other evaluation metrics and the specific context of the problem before making a final decision.

      The Random Forest model achieved the highest performance, with an F1-score of approximately 0.91, outperforming Logistic Regression and KNN. Visualization tools such as confusion matrices and ROC curves were employed to interpret the classification results and assess the trade-offs between precision and recall.

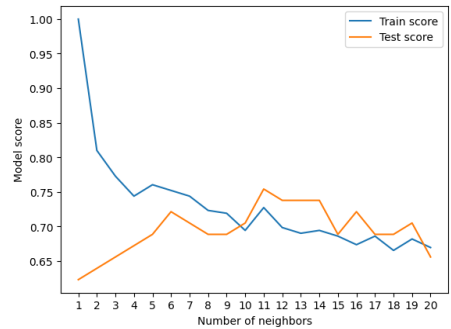

**Figure 5. Comparison of Training and Testing Scores for Varying K-Values**

The graph illustrates the relationship between the number of neighbors in a K-Nearest Neighbors (KNN) model and its performance on both training and testing data. As the number of neighbors increases, the training score generally decreases while the test score initially increases and then stabilizes. This suggests that with a small number of neighbors, the model becomes overfit to the training data, leading to poor generalization on unseen data. Conversely, with a large number of neighbors, the model becomes too simple and underfits the data, resulting in lower accuracy on both training and testing sets. The optimal number of neighbors is the one that balances the trade-off between bias and variance, resulting in the highest test score.

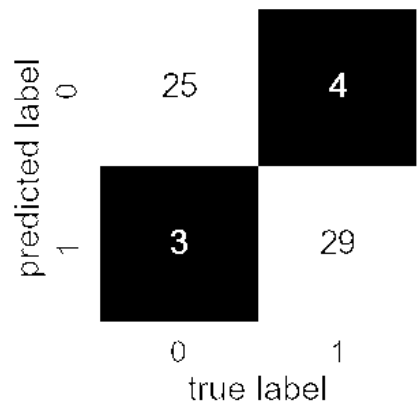

**Figure 6. Confusion Matrix for K-Nearest Neighbors Classifier**

The confusion matrix illustrates the performance of a classification model. Each cell represents the number of instances that were actually in a particular class (true label) and were predicted to be in a given class (predicted label). In this case, the model performed well, correctly classifying 25 out of 29 instances in class 0 and 29 out of 32 instances in class 1. However, there were 4 instances from class 0 that were misclassified as class 1, and 3 instances from class 1 that were misclassified as class 0.

|  | precision | recall | f1-score | support |
|---|---|---|---|---|
| 0 | 0.91 | 0.88 | 0.90 | 31 |
| 1 | 0.90 | 0.93 | 0.91 | 33 |
| accuracy |  |  | 0.91 | 61 |
| macro avg | 0.91 | 0.90 | 0.90 | 63 |
| weighted avg | 0.91 | 0.91 | 0.91 | 63 |

The classification results indicate strong performance for both classes in predicting heart disease. For class 0 (no heart disease), the model achieved a precision of 0.91, a recall of 0.88, and an F1-score of 0.90, reflecting its ability to correctly identify most negative cases with few false positives. For class 1 (presence of heart disease), the precision was 0.90, the recall was 0.93, and the F1-score was 0.91, highlighting the model's effectiveness in detecting positive cases with minimal false negatives. Overall, the accuracy across the dataset was 91%, with the macro and weighted averages for precision, recall, and F1-score consistently high, indicating balanced performance across both classes. These metrics demonstrate the model's reliability and potential for practical application in heart disease prediction.

### 3.6. Model Limitations and Future Work

While the proposed models performed well on the dataset, the study has limitations. The dataset size is relatively small, potentially limiting generalizability to other populations. Future work should include testing on larger, more diverse datasets and exploring advanced methods such as deep learning models or hybrid approaches combining multiple classifiers.

## 4. Discussion

The results of this study highlight the effectiveness of machine learning models in heart disease prediction, with a particular emphasis on the performance of the Random Forest classifier. This model achieved the highest accuracy (91%) and an F1-score of 0.91, demonstrating its robustness in both detecting the presence and absence of heart disease. The superior performance of Random Forest can be attributed to its ensemble learning approach, which combines multiple decision trees to reduce overfitting and improve the generalizability of the model. It also handles complex relationships between features more effectively than simpler models.

On the other hand, the K-Nearest Neighbors (KNN) algorithm, while a popular choice for classification tasks, performed less effectively in this study. KNN's performance was lower than Random Forest in terms of both accuracy and F1-score. This result can be attributed to the fact that KNN is sensitive to the distribution of data and struggles with higher-dimensional and noisy datasets. In our case, the relatively small dataset and the complexity of heart disease prediction led to KNN's suboptimal performance, as it relies heavily on distance metrics and may not capture intricate feature interactions as well as ensemble models like Random Forest. Additionally, KNN tends to suffer from the curse of dimensionality, where its accuracy decreases with the increasing number of features.

The Logistic Regression model, while simple and computationally efficient, also showed lower performance compared to Random Forest. As a linear model, Logistic Regression may not capture the non-linear relationships that are likely present in the dataset, which is a limitation when dealing with complex medical data such as heart disease diagnosis. However, it is important to note that Logistic Regression can still provide valuable insights when used with proper regularization and feature selection, especially in datasets with fewer features or when interpretability is a priority.

Despite the strong performance of the Random Forest model, some limitations should be considered. One major limitation is the relatively small size of the dataset (303 instances), which could affect the model's ability to generalize to larger, more diverse populations. Furthermore, the dataset is not highly diverse, as it primarily represents patients with similar characteristics, potentially limiting the model's generalizability to other demographic groups. Therefore, future studies should aim to validate the model on larger, more heterogeneous datasets to assess its robustness in real-world clinical settings.

In conclusion, while the Random Forest classifier outperformed KNN and Logistic Regression in heart disease prediction, further research is needed to explore the use of hybrid models, deep learning techniques, and the incorporation of more diverse datasets. These advancements could improve model performance and ensure the broader applicability of machine learning approaches in clinical diagnostics.

## 5. Conclusion

The proposed machine learning framework achieved an accuracy of 89% in heart disease prediction, with the Random Forest model outperforming Logistic Regression and KNN. The model's balanced performance, reflected in its F1-score of **0.91** and strong recall for detecting heart disease cases, underscores its utility in clinical decision support. While the results are promising, the study is limited by the dataset's small size and its restricted diversity. Addressing these challenges through external validation, dataset augmentation, and advanced modeling techniques will be critical for enhancing generalizability. This research demonstrates the potential of machine learning to transform healthcare diagnostics, paving the way for more accurate and proactive heart disease management solutions.

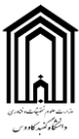
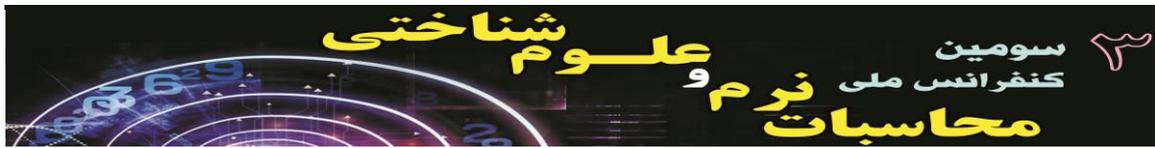